\documentclass[nohyperref]{article}

\usepackage[utf8x]{inputenc}

\usepackage{microtype}
\usepackage{graphicx}
\usepackage{booktabs} 

\usepackage{hyperref}

\usepackage[accepted]{icml2022}

\usepackage{subfigure}

\usepackage{amsmath}
\usepackage{amssymb}
\usepackage{mathtools}
\usepackage{amsthm}

\usepackage[capitalize,noabbrev]{cleveref}

\theoremstyle{plain}

\theoremstyle{definition}

\theoremstyle{remark}

\usepackage[textsize=tiny]{todonotes}
\usepackage{multirow}

\icmltitlerunning{Mixture of Input-Output Hidden Markov Models for Heterogeneous Disease Progression Modeling}

\begin{document}

\twocolumn[
\icmltitle{Mixture of Input-Output Hidden Markov Models \\for\\ Heterogeneous Disease Progression Modeling}




\begin{icmlauthorlist}
\icmlauthor{Taha Ceritli}{ox}
\icmlauthor{Andrew P.\ Creagh}{ox}
\icmlauthor{David A.\ Clifton}{ox}
\end{icmlauthorlist}

\icmlaffiliation{ox}{Department of Engineering Science, University of Oxford, Oxford OX3 7DQ, UK}

\icmlcorrespondingauthor{Taha Ceritli}{taha.ceritli@eng.ox.ac.uk}

\icmlkeywords{disease progression, hidden Markov models}

\vskip 0.3in
]

\printAffiliationsAndNotice{}  

\begin{abstract}
A particular challenge for disease progression modeling is the heterogeneity of a disease and its manifestations in the patients. Existing approaches often assume the presence of a single disease progression characteristics which is unlikely for neurodegenerative disorders such as Parkinson' disease. In this paper, we propose a hierarchical time-series model that can discover multiple disease progression dynamics. The proposed model is an extension of an input-output hidden Markov model that takes into account the clinical assessments of patients' health status and prescribed medications. We illustrate the benefits of our model using a synthetically generated dataset and a real-world longitudinal dataset for Parkinson's disease.  
\end{abstract}

\section{Introduction}
Disease progression refers to the temporal evolution of a disease over time. Modeling the temporal characteristics of a disease may be useful for various purposes including scientific discovery (e.g., understanding how a disease manifests itself by discovering the stages the patients typically go through) and clinical decision-making (e.g., evaluating the health status of a patient by identifying the stage the patient is in). 

Probabilistic time-series models are a natural choice for disease progression modeling as they take into account temporal relations in data. However, the task remains challenging for these models mainly because of (i) limited availability of data, (ii) data quality problems (e.g., missing data), (iii) the need for interpretability and (iv) heterogeneous nature of diseases such as Alzheimer's disease (AD) and Parkinson's disease (PD). A practical solution to these problems has been using hidden Markov models (HMMs), which (i) can be trained using small datasets, (ii) can handle missing data in a principled approach and (iii) are interpretable models, e.g., it is possible to relate inferred latent states to particular symptoms. Most existing HMMs \citep{jackson2003multistate,sukkar2012disease,guihenneuc2000modeling,wang2014unsupervised,sun2019probabilistic,severson2020personalized,severson2021discovery}, however, assume that each patient follows the same latent state transition dynamics, ignoring the heterogeneity in the disease progression dynamics.

The need for heterogeneous disease progression modeling has been highlighted by the works on \emph{disease subtyping}, which is defined as the task of identifying subpopulations of similar patients that can guide treatment decisions for a given individual \citep{saria2015subtyping}. Disease subtyping can be useful especially for complex diseases which are often poorly understood such as autism \citep{state2012emerging}, cardiovascular disease \citep{de2009heart} and Parkinson’s disease \citep{lewis2005heterogeneity}. The discovery of subtypes can further benefit both the scientific discovery (e.g., studying the associations between the shared characteristics of similar patients and potential causes) and clinical decision-making (e.g., reducing the uncertainty in an individual's expected outcome) \citep{saria2015subtyping}. 

Traditionally, disease subtyping has been carried out by clinicians who may notice the presence of subgroups \citep{barr1999patterns,ewing1921diffuse}. More recently, the growing availability of medical datasets and computational resources have facilitated the rapid adaptation of data-driven approaches that offer objective methods to discover underlying disease subtypes \citep{schulam2015clustering,lewis2005heterogeneity}. For instance, \citet{lewis2005heterogeneity} discover the presence of four subtypes of PD; however, they apply k-means clustering which may provide a limited capability to capture complex patterns in the data. \citet{schulam2015clustering} develop a more sophisticated approach based on a mixture model that is robust against the variability unrelated to disease subtpying; however, their proposed model does not take into account the temporal relations in the clinical visits. 

In this work, we relax the assumption of HMMs that the disease dynamics, as specified by the transition matrix, is shared among all patients. Instead, we propose the use of hierarchical HMMs for disease progression modeling, particularly mixture of HMMs (mHMMs) and their variants that can explicitly model group-level similarities of patients. We are motivated by the applications of mHMMs in other domains where they have been shown to outperform HMMs such as modeling activity levels in accelerometer data \citep{de2020mixture}, modeling clickstreams of web surfers \citep{ypma2008categorization} and modeling human mobility using geo-tagged social media data \citep{zhang2016gmove}. We summarize our contributions and the organization of the paper below:

\paragraph{Contributions:} To our knowledge, this is the first attempt to apply mHMMs to disease progression modeling. Particularly, we show that mixture of input-output HMMs (mIOHMMs) suits disease progression modeling better than IOHMMs, as they can discover multiple disease progression dynamics in addition taking into account the medications information. Moreover, we develop mixtures of a number of HMM variants, namely mIOHMMs, mixture of personalized HMMs (mPHMMs) and mixture of personalized IOHMMs (mPIOHMMs) which have not been explored before by the machine learning community.

\paragraph{Organization:} We first introduce our notation for HMMs and present three HMM variants with their mixture extensions (Section \ref{label:methodology}). We then discuss the related work (Section \ref{label:related-work}), which is followed by the experiments and the results (Section \ref{label:experiments}). Finally, we summarize our work and discuss the possible future research directions (Section \ref{label:summary}).

\section{Methodology}
\label{label:methodology}
This section describes the background information on HMMs, introduces our proposed models and the training procedure we apply. 

\subsection{Background}
Below we introduce our notation for HMMs and describe its three variants proposed by \citet{severson2020personalized}.

\subsubsection*{HMM}
We consider an HMM with a Gaussian observation model and define it as a tuple $M= (\pi, A, \mu, \Sigma)$ where $\pi$ is the initial-state probabilities, $A$ is the state-transition probabilities, $\mu$ and $\Sigma$ are the mean and covariance parameters of the observation model with Gaussian densities. The generative model of an HMM becomes as follows:

\begin{gather}
x_1^{(i)} \sim \mathcal{C}at(\pi), \qquad x_t^{(i)} | x_{t-1}^{(i)} = l \sim \mathcal{C}at(A_l), \nonumber \\
y_t^{(i)} | x_t^{(i)}=l \sim \mathcal{N}(y_t^{(i)}; \mu_l, \Sigma_l),
\end{gather}
where $x_t^{(i)}$ and $y_t^{(i)}$ are respectively the hidden state and observation at time $t$ for the $i^{th}$ time-series sequence, and $\mathcal{C}at(\cdot)$ and $\mathcal{N}(\cdot)$ respectively denote the Categorical and Gaussian distributions. Here, $x_{t}^{(i)}$ is conditionally generated given that the hidden state at time $t-1$ for the $i^{th}$ sequence, denoted by $x_{t-1}^{(i)}$, is the $l^{th}$ hidden state. Similarly, $y_{t}^{(i)}$ is generated conditionally on $x_{t}^{(i)}=l$.

\subsubsection*{PHMM}
We can train an HMM using multiple medical time-series sequences collected from different patients. This approach would rely on the assumption that each patient follows the same state means and covariance, which may not be realistic when the individuals differ from the state means with different amounts. To address this issue, \citet{severson2020personalized} propose a personalized HMM (PHMM) by modifying the observation model of HMM as follows:
\begin{eqnarray}
y_t^{(i)} | x_t^{(i)}=l \sim \mathcal{N}(y_t^{(i)}; \mu_l + r^{(i)}, \Sigma_l),
\end{eqnarray}
where $r^{(i)}$ denotes the individual deviation from the states.  

\subsubsection*{IOHMM}
The observed variables of an HMM are typically the clinical assessments made during hospital visits. However, the medications information can also be informative about the health status of a patient. To incorporate such information into the disease progression modeling, \citet{severson2020personalized} introduce the following observation model:
\begin{eqnarray}
\label{eqn:iohmm-likelihood}
y_t^{(i)} | x_t^{(i)}=l &\sim&  \mathcal{N}(y_t^{(i)}; \mu_l +v_l d_t^{(i)}, \Sigma_l),
\end{eqnarray}
where $d_t^{(i)}$ is the observed medication data at time $t$ for the $i^{th}$ patient and $v_l$ denotes the state medication effects.  

The proposed model resembles input-output HMMs \cite{bengio1994input} except that the hidden states are not conditioned on the input variables which are used to incorporate medications data, as the medication is thought to have no disease-modifying impact. This assumption is valid for diseases such as PD where there is no cure but treatments that can only help reduce the symptoms.

\subsubsection*{PIOHMM}
Finally, combining PHMM and IOHMM provides us a personalized model that takes into account the medications:
\begin{eqnarray}
y_t^{(i)} | x_t^{(i)}=l &\sim&  \mathcal{N}(y_t^{(i)}; \mu_l + r^{(i)} + (v_l + m^{(i)}) d_t^{(i)}, \Sigma_l),
\end{eqnarray}
where $m^{(i)}$ denotes the personalized medication effects.

\subsection{The Proposed Models}
Below we follow a general recipe to construct hierarchical mixture models. We first extend the HMM and then its three variants to their mixture correspondences. 

We construct the mixture version of a HMM model (e.g., mPHMMs) by concatenating the parameters of each HMM variant (e.g., PHMM) for simplicity; however, it would also be possible to apply alternative schemes, e.g., see \citet{smyth1996clustering} for a hierarchical clustering-based approach. 

\subsubsection*{mHMMs}
\begin{figure}[htb!]
\begin{center}
\includegraphics[width=0.45\textwidth]{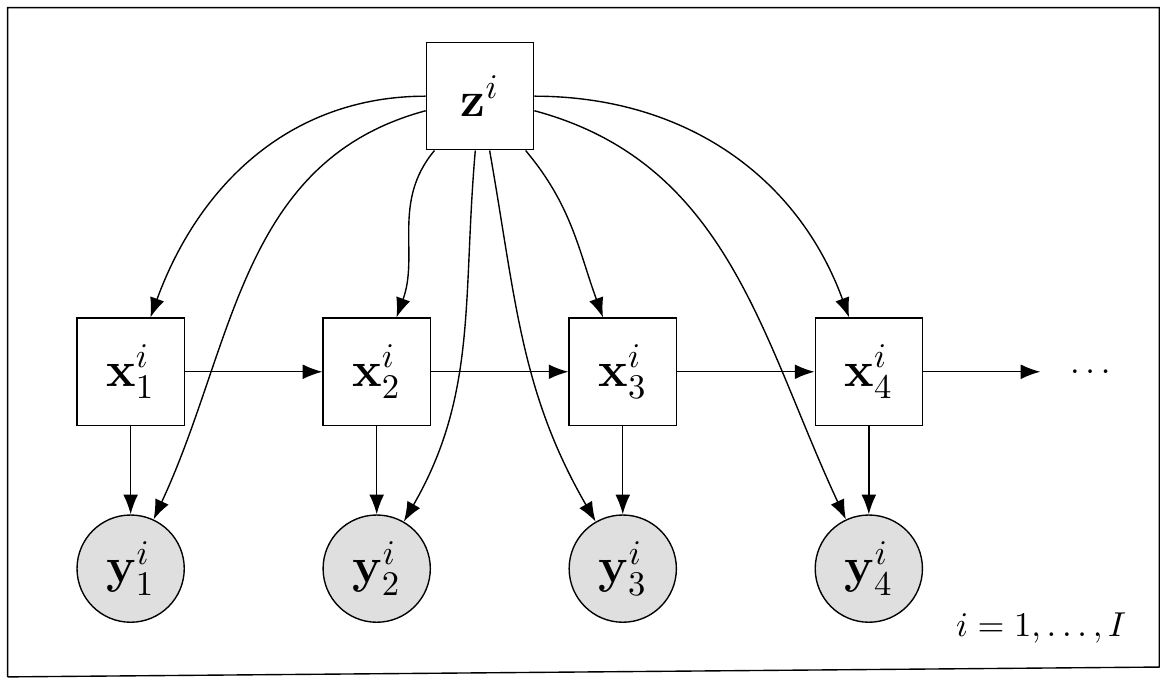}
\caption{A graphical representation of mHMMs.}
\label{fig:mHMMs-graphical}
\end{center}
\end{figure}

We define a mHMMs as a set $M= \{M_1, M_2, \dots, M_K\}$ where $M_k = (\pi_k, A_k, \mu_k, \Sigma_k)$
is the $k^{th}$ HMM mixture. The generative model becomes as follows:

\begin{align}
z^{(i)} &\sim \mathcal{C}at(\alpha), \nonumber \\
x_1^{(i)} | z^{(i)}=k &\sim \mathcal{C}at(\pi_k), \nonumber \\
x_t^{(i)} | x_{t-1}^{(i)} = l, z^{(i)}=k &\sim \mathcal{C}at(A_{k,l}), \nonumber \\
y_t^{(i)} | x_t^{(i)}=l, z^{(i)}=k &\sim \mathcal{N}(y_t^{(i)}; \mu_{k,l}, \Sigma_{k,l}),
\end{align}
where $z^{(i)}$ denotes the HMM that the $i^{th}$ time-series sequence belongs to, $x_t^{(i)}$ and $y_t^{(i)}$ are respectively the corresponding hidden state and observation at time $t$. Note that when the cardinality of $z$ is 1, the model reduces to the standard HMM. Fig.\ \ref{fig:mHMMs-graphical} presents a graphical representation of mHMMs.

mHMMs assume that each time-series sequence belongs to an HMM mixture. This construction allows us to cluster similar sequences so that each cluster is represented using different parameter values. As we have mentioned earlier, training a single HMM for all sequences may not be expressive enough. On the other hand, training a separate HMM for each sequence can be challenging due to the sparsity of the data and the computational problems. mHMMs overcome these problems by combining a number of HMMs which is higher than 1 and lower than the number of sequences. 

\subsubsection*{mPHMMs}
Similarly to mHMMs, we obtain the mixture versions of the HMM variants. For example, we modify the observation model of PHMM to obtain its mixture version as follows:
\begin{eqnarray}
y_t^{(i)} | x_t^{(i)}=l, z^{(i)}=k \sim \mathcal{N}(y_t^{(i)}; \mu_{k,l} + r^{(i)}, \Sigma_{k,l}).
\end{eqnarray}

\subsubsection*{mIOHMMs}
We obtain mIOHMMs using the observation model given below:
\begin{eqnarray}
\label{eqn:miohmm-likelihood}
y_t^{(i)} | x_t^{(i)}=l, z^{(i)}=k \sim \mathcal{N}(y_t^{(i)}; \mu_{k,l} + v_{k,l} d_t^{(i)}, \Sigma_{k,l}).
\end{eqnarray}

\subsubsection*{mPIOHMM}
Finally, mPIOHMM has the following observation model:
\begin{eqnarray}
y_t^{(i)} \sim  \mathcal{N}(y_t^{(i)}; \mu_{k,l} + r^{(i)} + (v_{k,l} + m^{(i)}) d_t^{(i)}, \Sigma_{k,l}),
\end{eqnarray}
where $l=x_t^{(i)}$ and $k=z^{(i)}$.

\subsection{The Training of the Models}
We follow the same training procedure proposed by \citet{severson2020personalized} where variational inference is used to approximate the posterior distributions over the latent variables of $x$, $m$ and $r$ as follows:

\begin{align}
q(x,m,r|y, \lambda) &=  \prod_{i=1}^N q(m^{(i)}|\lambda) q(r^{(i)}|\lambda) q(x^{(i)}| y^{(i)}, m^{(i)}, r^{(i)}), \nonumber \\
&=  \prod_{i=1}^N q(m^{(i)}|\lambda) q(r^{(i)}|\lambda) \nonumber \\
& \qquad \qquad \prod_{t=2}^{T_i} q(x^{(i)}_t|x^{(i)}_{t-1},y^{(i)}_t, m^{(i)}, r^{(i)}),
\end{align}
where $\lambda$ denote the variational free parameters. The corresponding evidence lower bound (ELBO) is maximized using coordinate ascent alternating between the updates for variational parameters $\lambda$ and model parameters $\theta$. Please see \citet{severson2020personalized} for the details of the training algorithm. Note that we simplify the inference by not explicitly inferring the latent variables $z$. Instead, we obtain the cluster membership of each sequence based on its state trajectory estimated via the Viterbi algorithm, thanks to the block-diagonal structure of the transition matrices. However, it would be possible to explicitly infer the variables $z$ by introducing the corresponding variational distribution $q(z_i | \lambda_{z_i})$.

\section{Related Work}
\label{label:related-work}
Most common approach for disease progression modeling has been using HMMs. For example, \citet{guihenneuc2000modeling} employ a HMM with discrete observations for modeling the progression of Acquired Immune Deficiency Syndrome (AIDS). \citet{sukkar2012disease} apply the same model for Alzheimer's Disease. \citet{wang2014unsupervised} introduce additional hidden variables to incorporate the \emph{comorbidities} of a disease into the transition dynamics. Note that comorbidities are defined as syndromes co-occurring with the target disease, e.g., hypertension is a common comorbidity of diabetes. Other applications of HMMs for disease progression include the work on Huntington’s disease \cite{sun2019probabilistic} and abdominal aortic aneurysm \cite{jackson2003multistate}. Lastly, the standard HMMs have been modified for personalized disease progression. \citet{altman2007mixed} introduce random effects to better capture individual deviations from states. \citet{severson2020personalized,severson2021discovery} propose a model that is both personalized and takes into account medications information for disease progression modeling. 

An alternative approach to personalize disease progression is through Gaussian processes (GPs). \citet{peterson2017personalized} propose a GP model personalized based on each patient’s previous visits. \citet{lorenzi2019probabilistic} combines a GP with a set of random effect variables where the former is used to model progression dynamics shared among patients and the latter is used to represent their individual differences. \citet{schulam2015framework} propose a more general framework based on a hierarchical GP model with population, subpopulation and individual components that has been applied on the measurements of a single biomarker. \citet{futoma2016predicting} later generalize this model to the case of multiple biomarkers. 

Another common approach for disease progression modeling have been the use of deep learning, especially when the interpretability is not a major concern and a large amount of clinical data is available \citep{che2018hierarchical,eulenberg2017reconstructing,pham2017predicting,alaa2019attentive,lee2020temporal,chen2022clustering}. Among these methods, the most relevant works to ours are the approaches proposed by \citet{lee2020temporal} and \citet{chen2022clustering} which can identify ``similar'' patients via time-series clustering. 

Perhaps the closest related works are the studies on disease subtyping, particularly those focusing on Parkinson's Disease (PD). \citet{lewis2005heterogeneity} discover the presence of four subtypes of PD by applying k-means clustering. \citet{schulam2015clustering} develop a mixture model that is robust against the variability unrelated to disease subtpying. Both these approaches, however, do not take into account the temporal relations in the clinical visits. 

Finally, mHMMs have been shown to outperform HMMs in other domains such as modeling activity levels in accelerometer data \citep{de2020mixture}, modeling clickstreams of web surfers \citep{ypma2008categorization} and modeling human mobility using geo-tagged social media data \citep{zhang2016gmove}. We also note a couple of works on the training of mHMMs such as the hierarchical clustering-based approach proposed by \citet{smyth1996clustering} and the spectral-learning based training algorithm proposed by \citet{subakan2014spectral}.

\section{Experiments} 
\label{label:experiments}

We present two sets of experiments. The goal of the first experiment is to demonstrate the ability of mPHMM to simultaneously learn personalized state effects and multiple disease progression dynamics using synthetically generated data, for which we know the true disease progression dynamics. Then, we show that mIOHMM provides a better `fit' of a real-world dataset than IOHMM by discovering multiple disease progression dynamics. The code to reproduce the experiments is publicly available at \url{https://github.com/tahaceritli/mIOHMM}.

\begin{figure*}[htb!]
\begin{center}
\includegraphics[width=0.9\textwidth]{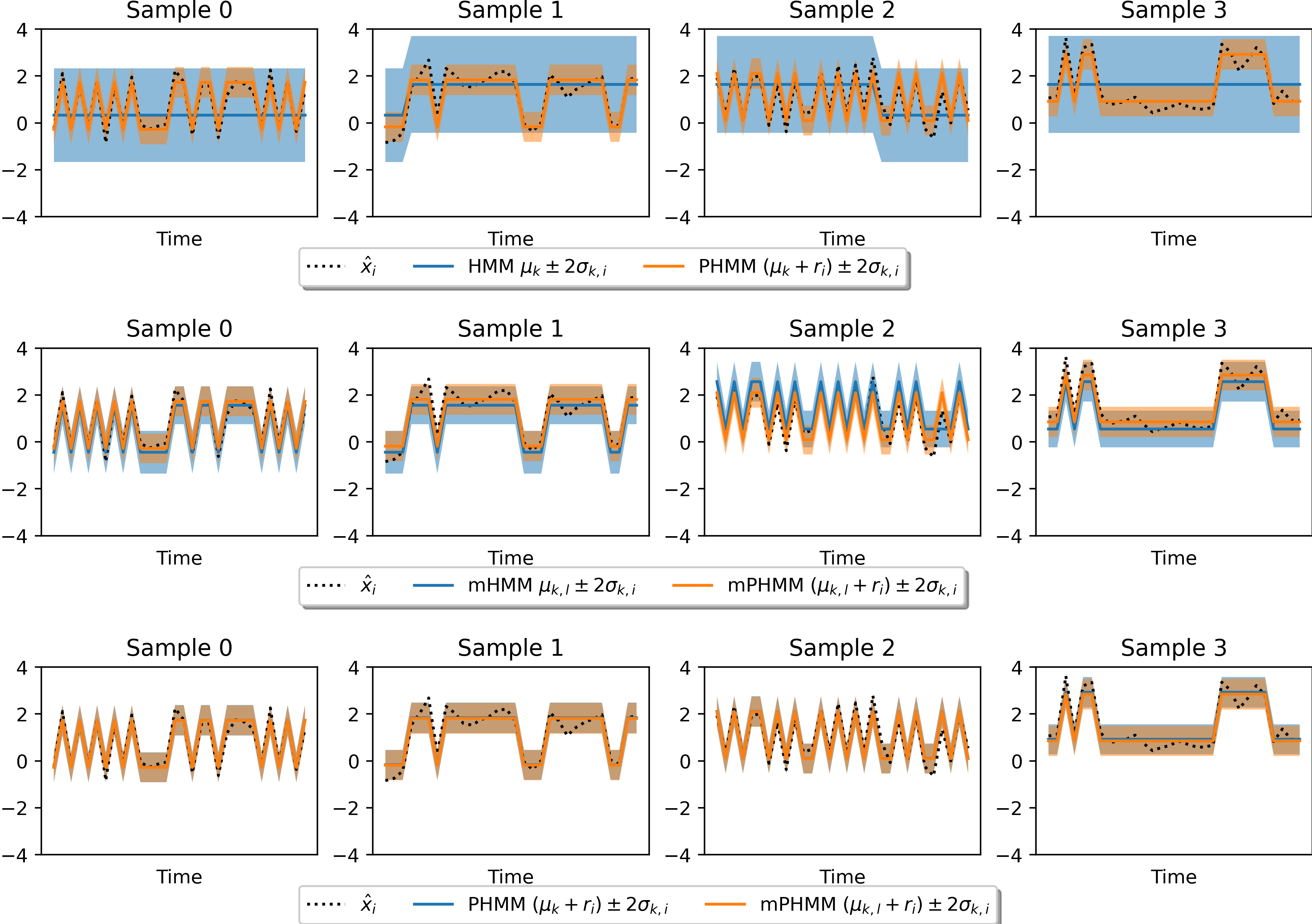}
\caption{A comparison of the models for simulated data based on the original study by \citet{severson2020personalized}. The three rows correspond to different pairs of models being compared. The standard HMM incorrectly assigns states and compensates for the personalization with large variances, as shown in the first row. We observe the same phenomenon in the middle row with mHMMs, although the variance is lower than the variance of HMM as the mixture components provide a richer representation of the state-means. As per the bottom comparison, PHMM and mPHMMs overlap showing that the model can still handle individual variations in the data.}
\label{fig:synt-fit-plot}
\end{center}
\end{figure*}

\subsection{Synthetic Data}
Combining the settings used by \citet{severson2020personalized} and \citet{smyth1996clustering}, we build a 2-component mPHMM with 2 latent states for each PHMM. The state transition matrices of the PHMM mixtures are given below:
\begin{equation*}
A_1 = 
\begin{bmatrix}
0.8 & 0.2  \\
0.2 & 0.8
\end{bmatrix},
A_2 = 
\begin{bmatrix}
0.2 & 0.8  \\
0.8 & 0.2
\end{bmatrix},
\end{equation*}
where $A_k$ denotes the state transition matrix of the $k^{th}$ PHMM. The observation model is built using Gaussian densities with the means $\mu_1 = \mu_2 =
\begin{bmatrix}
0 \\ 
2 
\end{bmatrix}$ and variances $\sigma_1^2 = \sigma_2^2 =
\begin{bmatrix}
0.1 \\
0.1 
\end{bmatrix}$. Note that the state means and variances are the same for each PHMM whereas the transition dynamics are different, i.e., the transitions between the latent states are more likely to occur in the first PHMM than they are in the second PHMM. The initial state probabilities are assumed to be uniformly distributed. We use the noisy observation model of $\hat{x}_{i,t} | z_{i,t} = l \sim \mathcal{N}(\mu_l + r_i, \Sigma_T)$ where $\Sigma_T$ is specified via a squared exponential kernel $\kappa(t, t′) = \sigma^2 \exp \frac{-(t-t')^2}{2*l^2}$ with $l$ and $\sigma$ are respectively set to 1 and 0.1. Lastly, the personalized state offset $r_i$ is uniformly sampled for each sample with $b=1$.

Fixing the dimensionality of the data to 1, we generate 200 sequences of length 30 using this model. The training of mPHMM yields the parameter estimates given below: 
\begin{align*}
\hat{A_1}= 
\begin{bmatrix}
0.80 & 0.20  \\
0.19 & 0.81
\end{bmatrix},
\hspace{.5cm}
\hat{\mu_1} = \begin{bmatrix}
0.11 \\
2.10 
\end{bmatrix},
\hspace{.5cm}
\hat{\sigma^2_1} = 
\begin{bmatrix}
0.10 \\
0.11
\end{bmatrix},
\\
\hat{A_2} = 
\begin{bmatrix}
0.21 & 0.79  \\
0.80 & 0.20
\end{bmatrix},
\hspace{.5cm}
\hat{\mu_2} = \begin{bmatrix}
0.04 \\
2.05 
\end{bmatrix},
\hspace{.5cm}
\hat{\sigma^2_2} = 
\begin{bmatrix}
0.10 \\ 
0.10 
\end{bmatrix},
\end{align*}

On the other hand, we obtain the following parameter estimates using PHMM: $A = \begin{bmatrix}
0.53 & 0.47  \\
0.46 & 0.54
\end{bmatrix}$, $\mu = \begin{bmatrix}
0.05 & 2.05 
\end{bmatrix}$ and $\sigma = \begin{bmatrix}
0.10 & 0.11 
\end{bmatrix}$ which indicates that PHMM cannot distinguish the heterogeneous state-transitions. Note that we could have adapted PHMM to this example by using 4 latent states; however, the distinction between the states would not be clear as the block-diagonal structure is not introduced in PHMM (see the additional experimental results in Appendix \ref{appendix:additional-experimental-results}). 

Finally, we demonstrate that our model keeps the personalization capabilities of the original PHMM discussed in \citet{severson2020personalized}. Fig.\ \ref{fig:synt-fit-plot} presents a number of sequences and the corresponding estimates obtained using HMM, PHMM, mHMM and mPHMM. The figure indicates that mPHMM performs similarly as PHMM in fitting the data. However, mPHMM has the advantage of discovering the heterogeneous transition matrices over PHMM as discussed above.

\subsection{Real Data}
\subsubsection*{Data}
Following the experimental setup in \citet{severson2020personalized}, we use the Parkinson Progression Marker Initiative (PPMI) dataset \cite{marek2011parkinson} for real data experiments. PPMI is a longitudinal dataset collected from 423 PD patients, including clinical, imaging and biospecimen information. We focus on the clinical assessments measured via the Movement Disorder Society Unified Parkinson’s Disease Rating Scale (MDS-UPDRS) \cite{goetz2008movement}. The MDS-UPDRS consists of a combination of patient reported-measures and physician-assessed measures: (i) non-motor experiences of daily living, (ii) motor experiences of daily living, (iii) motor examination and (iv) motor complications. Each item on the scale is rated from 0 (normal) to 4 (severe). We do not use the motor complications obtaining 59 features for the observations.

As the medication data, we use the levodopa equivalent daily dose (LEDD) \cite{tomlinson2010systematic}, which is provided in the PPMI dataset.

\subsubsection*{Metrics}
To compare the models, we use three information criteria: Akaike Information Criterion (AIC, \citealt{akaike1998information}), Bayesian Information Criterion (BIC, \citealt{schwarz1978estimating}) and Integrated Completed Likelihood (ICL, \citealt{biernacki2000assessing}) which are defined below:
\begin{eqnarray*}
\mathrm{AIC} &=& - 2\ell + 2k \\
\mathrm{BIC} &=& - 2\ell + k \log N \\
\mathrm{ICL} &=& - 2\hat{\ell} + 2k, 
\end{eqnarray*}
where $\ell$ is the log-likelihood of the training data, $k$ is the number of free parameters, $N$ is the number of training data instances, and $\hat{\ell}$ is the log-likelihood of the training data under the most likely trajectory. Here, $k$ is calculated as $L^2 + 3*L*D -1$ where $D$ is the dimension of observations and $L$ is the total number of hidden states aggregated over the HMM mixtures, as we use diagonal covariance matrices. Additionally, the log-likelihoods are calculated based on Equations \ref{eqn:iohmm-likelihood} and \ref{eqn:miohmm-likelihood}.

\subsubsection*{Model}
We compare IOHMM and a number of mIOHMMs with varying number of components (i.e., $K \in \{2,3,4,5\}$). Note that these models are not personalized, meaning that they are equivalent to PIOHMM with personalized effect variables fixed to zero, i.e., $r_i=0$ and $m_i=0$. In this work, we evaluate the impact of using mIOHMM over IOHMM. One could similarly apply model selection for PIOHMM without fixing personalized effect variables to zero; however, in our experience, this is computationally expensive and more efficient algorithms need to be developed which is out of the scope of this work. Additionally, we fix the number of hidden states to 8 following the setting in \citet{severson2020personalized}, and use diagonal covariance matrices.

\subsubsection*{Results}
Table \ref{tab:all-methods-performance} presents the values of the information criteria obtained using the models, which indicate that mIOHMMs are favoured over IOHMM. As per the table, AIC and BIC select five components whereas ICL selects two components. This result is not surprising as AIC and BIC tend to be overoptimistic about the model size \cite{biernacki2000assessing}.  

\begin{table}[h]
\centering
\caption{Performance of the mHMM methods using AIC, BIC and ICL.}
\begin{tabular}{lccc}
\toprule
K &  AIC &  BIC & ICL \\
\midrule
1 & -5.5370e+07 & -5.5365e+07 & -5.4256e+07  \\
2 & -5.5532e+07 & -5.5520e+07 & \textbf{-5.5330e+07} \\
3 & -5.5540e+07 &	-5.5521e+07 & -5.5246e+07 \\
4 & \textbf{-5.5567e+07} & \textbf{-5.5542e+07} & -5.2234e+07 \\
5 & -5.5536e+07 & -5.5503e+07 & -5.5250e+07 \\
\bottomrule
\end{tabular}
\label{tab:all-methods-performance}
\end{table}

Following the ICL criterion, we compare and interpret the parameter estimates obtained using mIOHMMs with 2
components and IOHMM. In addition to the ICL criterion that reflects the overall performance of the models, we report a measure of performance per patient in Appendix Fig.\ \ref{fig:appendix:ppmi-test-diff} which indicates that mIOHMM leads to a higher likelihood per patient than IOHMM (on average). Next, we inspect the initial-state probabilities, transition-state probabilities, state-means and medication-means.

\begin{figure*}[t!]
\centering     
\subfigure[IOHMM]{\includegraphics[width=.95\textwidth]{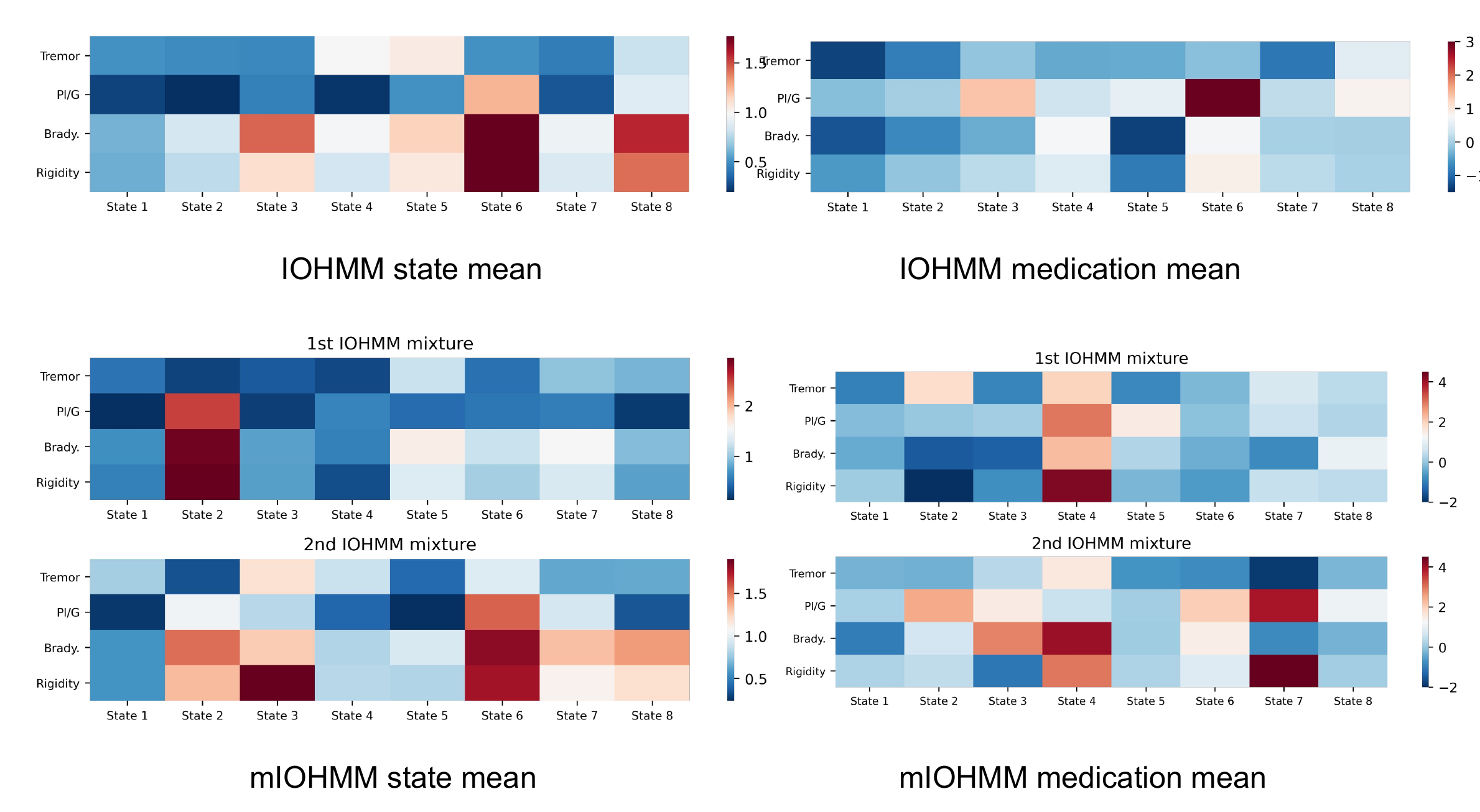}}
\caption{A summary of the state and medication means obtained using IOHMM and 2-component mIOHMM.}
\label{fig:mean-summaries}
\end{figure*}

Note that the two clusters obtained using mIOHMMs respectively contain 105 and 227 patients. Table \ref{tab:summary-patients-char} presents a summary of the age and sex distribution for each cluster of patients, which indicates that the clusters are not picking up on simple subject demographics.

\begin{table}[h]
\centering
\caption{A summary of the patients' characteristics.}
\begin{tabular}{llcrr}
\toprule
 & & Overall &  1st cluster & 2nd cluster \\
\midrule
Age &  & 61.6 (9.8) & 60.6 (10.1) & 62.1 (9.6)\\
\multirow{2}{*}{Sex} & Female & 217 (65\%) & 76 (72\%) & 141 (62\%)\\
 & Male & 115 (35\%) & 29 (28\%) & 86 (38\%)\\
\bottomrule
\end{tabular}
\label{tab:summary-patients-char}
\end{table}

We have a total number of 59 features. Therefore, the complete state and medication means are reported in Appendix \ref{appendix:additional-experimental-results}. Here, we present their summaries which are calculated based on primary clinical symptoms used for the diagnoses of PD as done in \citet{severson2020personalized}. Fig.\ \ref{fig:mean-summaries} presents the average of state and medication for each hidden state based on the tremor, bradykinesia, rigidity and postural instability/gait (PI/G) related features. The relevant features are selected based on the MDS-UPDRS as follows: tremor, 2.10, 3.15-3.18; postural instability gait, 2.12-2.13, 3.10-3.12; bradykinesia, 3.4-3.8, 3.14; and rigidity, 3.3 (see \citet{stebbins2013identify} for the details).

We first discuss the initial-state probabilities. Recall that the state-transitions are allowed only in the forward direction. Therefore, we expect the most likely initial-states to represent a patient's health condition at enrollment, which is often mild. This is indeed the case for both IOHMM and each mixture of mIOHMMs where the most likely initial-states have mild symptoms. For instance, the state 2 of IOHMM has the highest initial-state probability and the lowest total MDS-UPDRS score. For the first and second mixtures of mIOHMMs, these are respectively the states 1 and 5.

Note that the score for each symptom is not recommended to be collapsed into a single total score \cite{goetz2008movement}. Therefore, we also consider the score per symptom and discuss the characteristics of the states based on the intensities of individual scores.

One common state characteristics is the co-occurring severity in the PI/G, bradykinesia and rigidity symptoms. For example, the state 6 of IOHMM, the state 2 of the first mixture of mIOHMMs and the state 6 of the second mixture of mIOHMMs have severe PI/G, bradykinesia and rigidity issues but no tremor issues. However, these states differ in terms of the level of severity, e.g., the state with least severe symptoms is the state 2 of IOHMM. 

Another state characteristics we observe is the co-occurring severity in bradykinesia and rigidity symptoms. This characteristics is seen in the states 3 and 8 of IOHMM, and the states 2 and 8 of the second mixture of mIOHMMs.

We characterize the states based on the subtype methodology proposed by \citet{stebbins2013identify}, where each state is labeled with one of the following subtypes: (i) tremor, (ii) PI/G and (iii) intermediate. For IOHMM, the states 1,2,4,5,7 are tremor dominant, states 6 is PI/G dominant, and states 3 and 8 are indeterminate. For the first mixture of mIOHMMs, the states 1,3,5,7,8 are tremor dominant, states 2 and 4 are PI/G dominant and state 7 is indeterminate. For the second mixture of mIOHMMs, the states 2,6,7 are PI/G dominant and the remaining states are tremor dominant. These observations conform with the findings of \citet{severson2020personalized} in that the number of tremor dominant states is higher than the number of PI/G dominant states.

\begin{figure*}[t!]
\begin{center}
\includegraphics[width=.7\textwidth]{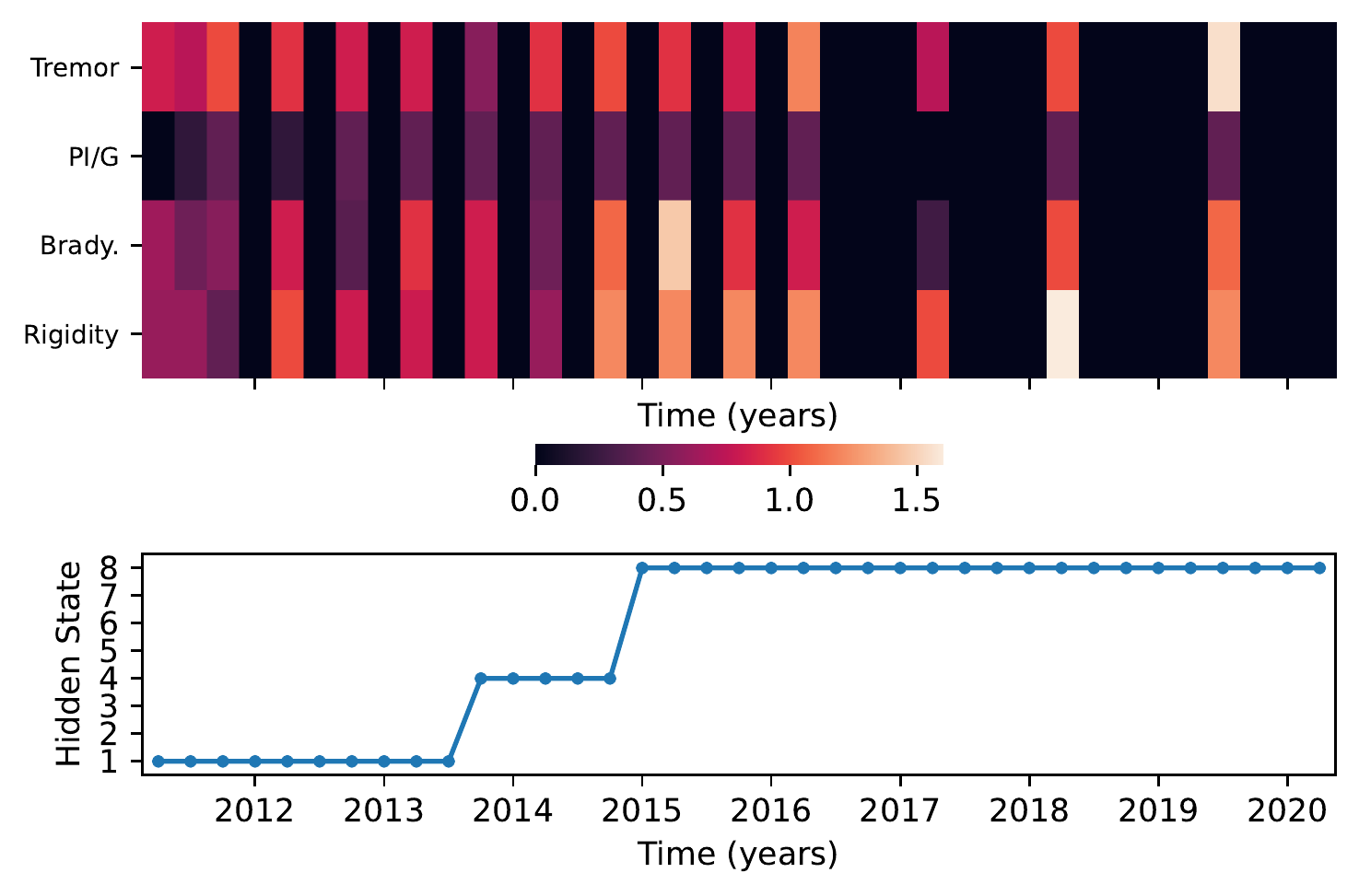}
\caption{A histogram of a deteriorating PD patient (female, 57 years old) (Top) and the corresponding state-trajectory (Bottom). The data is clustered into the second mixture of mIOHMMs. The subject appears to deteriorate over time, as denoted by visiting the states 1, 4 and 8---indicating increasing means in the bradykinesia- and rigidity-based features. Notice that the data column is entirely missing when the patient misses a hospital visit, whereas some features are non-missing but still zero-valued because a 0 rating has been given for the corresponding symptom.}
\label{fig:data-state-traj}
\end{center}
\end{figure*}

In addition to the state-means, each state is associated with a medication variable. When a patient is on medication, the symptoms are modeled using the state means and state medication variables. This can help to distinguish the states with similar means and different medications effects. For example, the state 5 and 7 of the first mixture of mIOHMMs are similar in terms of the state-means; however, the state 5 has a higher medication effect for the PI/G symptom than the state 5. Note that one should take into account whether a patient is on medication at a hospital visit and, if so, the dose of the medication for a better understanding of the model findings. 

We note that the state-transition probabilities favour self-transitions which may not be surprising as the disease progression occurs slowly between hospital visits. Again, the transitions need to be interpreted considering the medications effects.

Finally, we discuss the state-trajectories obtained via the Viterbi algorithm. Fig.\ \ref{fig:data-state-traj} plots a patient's data who is clustered in the second mIOHMM mixture and the corresponding state-trajectory. It was observed the changes in the regime have been successfully captured by the states. Note: for the corresponding feature list and index, we refer the reader to Appendix Fig.\ \ref{fig:appendix:IOHMM-mIOHMM-full-state-means}. We also visualize the disease progression trajectories for a number of patients in each cluster with their overall severity scores in Appendix Fig.\ \ref{fig:appendix:state-traj}

\section{Summary} 
\label{label:summary}
In this paper, we have applied mixtures of hidden Markov models for disease progression modeling. The proposed models can identify similar groups of patients through time-series clustering and separately represent the progression of each group, unlike hidden Markov models which assume that a single dynamics is shared among all patients. Our experiments on a real-world dataset have demonstrated the benefits of mixture models over a single hidden Markov model for disease progression modeling. Future work includes the development of efficient training algorithms for mPIOHMMs. 

\section*{Acknowledgements}
We would like to thank Kristen A.\ Severson for comments and discussion. Data used in the preparation of this article were obtained from the Parkinson’s Progression Markers Initiative (PPMI) database (\url{www.ppmi-info.org/access-data-specimens/download-data}). For up-to-date information on the study, visit \url{www.ppmi-info.org}. PPMI – a public-private partnership – is funded by the
Michael J. Fox Foundation for Parkinson’s Research and funding partners, including 4D Pharma, Abbvie, AcureX Therapeutics, Allergan, Amathus Therapeutics, Aligning Science Across Parkinson’s, Avid Radiopharmaceuticals, Bial Biotech, Biogen, BioLegend, Bristol-Myers Squibb, Calico Life Sciences LLC, Celgene, DaCapo Brainscience, Denali, the Edmond J. Safra Foundation, Eli Lilly and Company, GE Healthcare, Genentech, GlaxoSmithKline, Golub Capital, Handl Therapeutics, Insitro, Janssen, Lundbeck, Merck, Meso Scale Diagnostics, Neurocrine Biosciences, Pfizer, Primal, Prevail Therapeutics, Roche, Sanofi Genzyme, Servier, Takeda, Teva, UCB, Vanqua Bio, Verily, Voyager Therapeutics and Yumanity.

\bibliography{references}
\bibliographystyle{icml2022}

\newpage
\appendix
\onecolumn

\section*{Appendices}

\section{Additional Experimental Results}
\label{appendix:additional-experimental-results}

\subsection{Synthetic Data}
The training of PHMM with 4 latent states yields the parameter estimates given below:
\begin{align*}
\hat{A}= 
\begin{bmatrix}
0.55 & 0.39 & 0.09 & 0.06 \\
0.36 & 0.48 & 0.08 & 0.08 \\
0.11 & 0.11 & 0.36 & 0.43 \\
0.08 & 0.11 & 0.42 & 0.39
\end{bmatrix},
\hspace{.5cm}
\hat{\mu} = \begin{bmatrix}
-0.15 \\
1.85 \\
0.32 \\
2.34 
\end{bmatrix},
\hspace{.5cm}
\hat{\sigma^2} = 
\begin{bmatrix}
0.06 \\
0.05 \\
0.05 \\
0.05 
\end{bmatrix},
\end{align*}
where $\hat{A}$ denotes the estimate for the state transition matrix, and $\hat{\mu}$ and $\hat{\sigma^2}$ are the estimates for the state means and variances.

\subsection{Real Data}

\begin{figure}[h]
\begin{center}
\includegraphics[width=.4\textwidth]{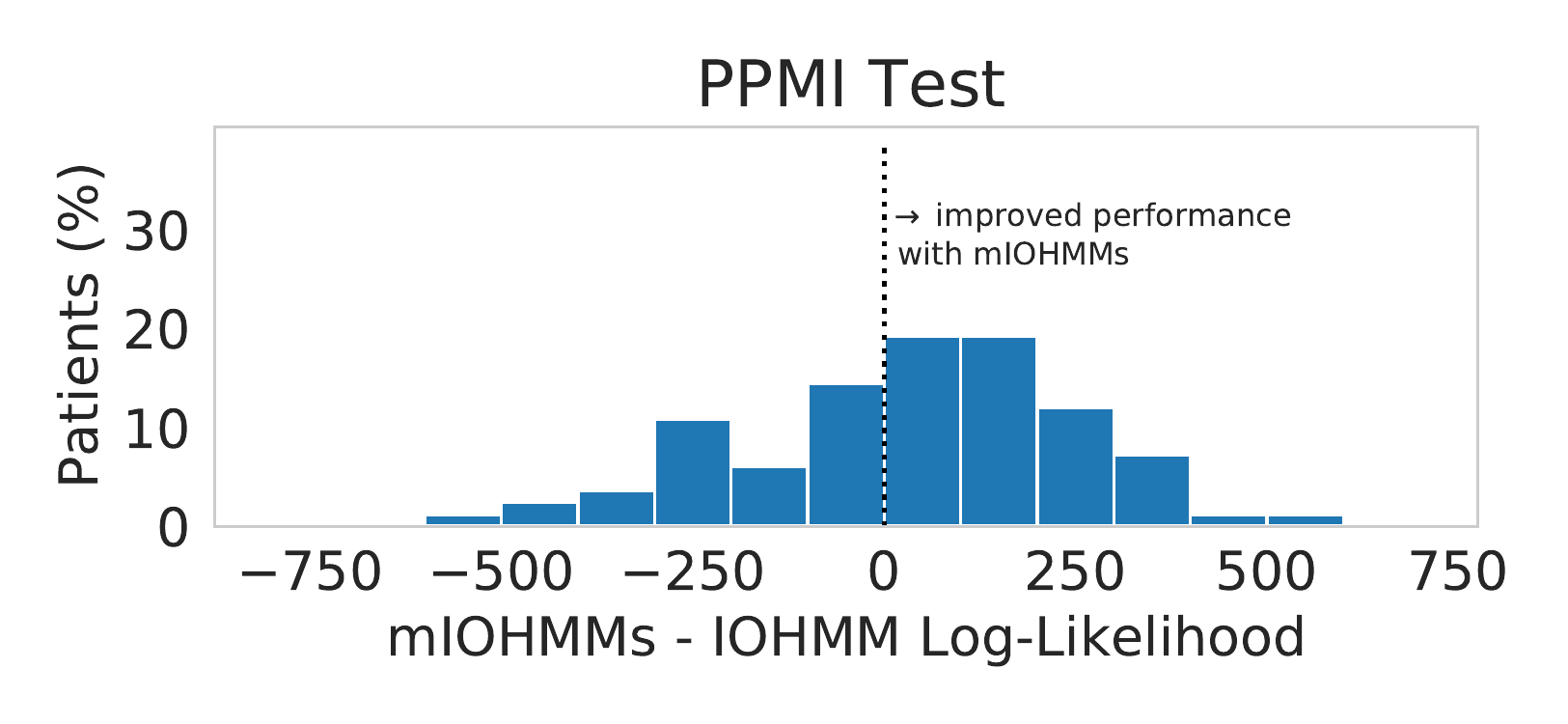}
\caption{Difference in the log-likelihood of the models for each patient in the test data, plotted with respect to the percentage of patients.}
\label{fig:appendix:ppmi-test-diff}
\end{center}
\end{figure}

\begin{figure*}[htb!]
\centering     
\subfigure[IOHMM]{\label{fig:a}\includegraphics[width=.2\textwidth]{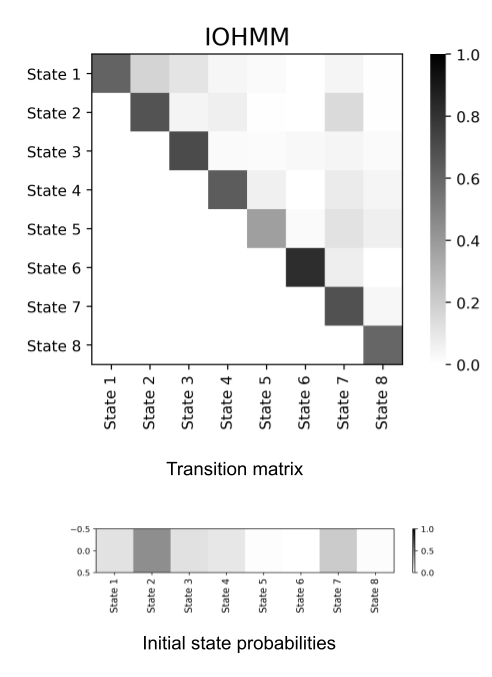}}\hspace{1cm}
\subfigure[mIOHMM]{\label{fig:b}\includegraphics[width=.4\textwidth]{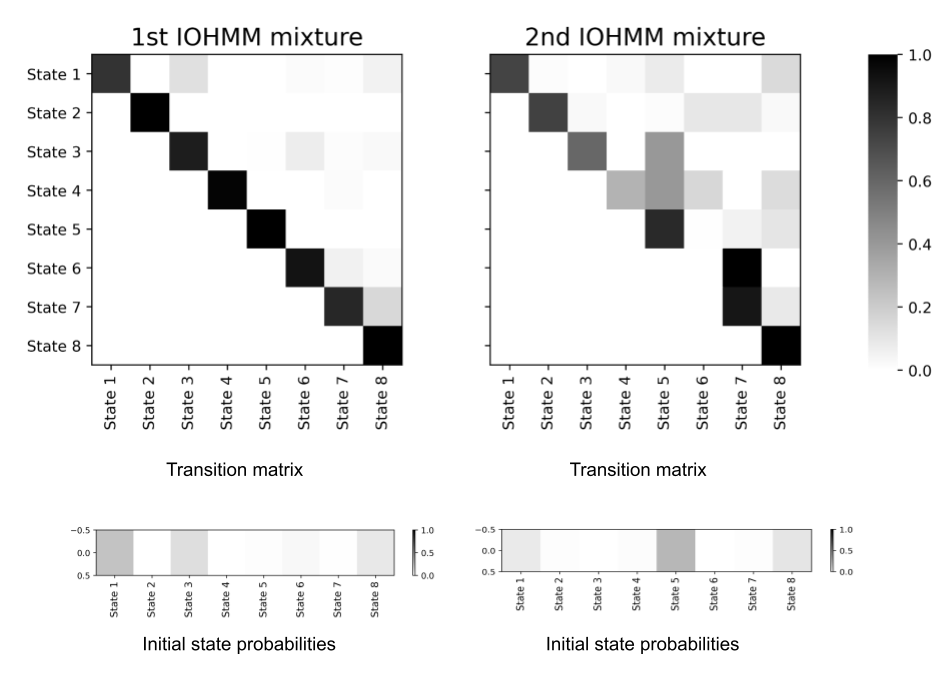}}
\caption{Initial-state probabilities and transition matrices obtained using IOHMM and mIOHMMs with 2 mixtures.}
\label{fig:appendix:init-tran-params-models}
\end{figure*}

\begin{figure*}[htb!]
\centering     
\subfigure[IOHMM]{\label{fig:a}\includegraphics[width=.7\textwidth]{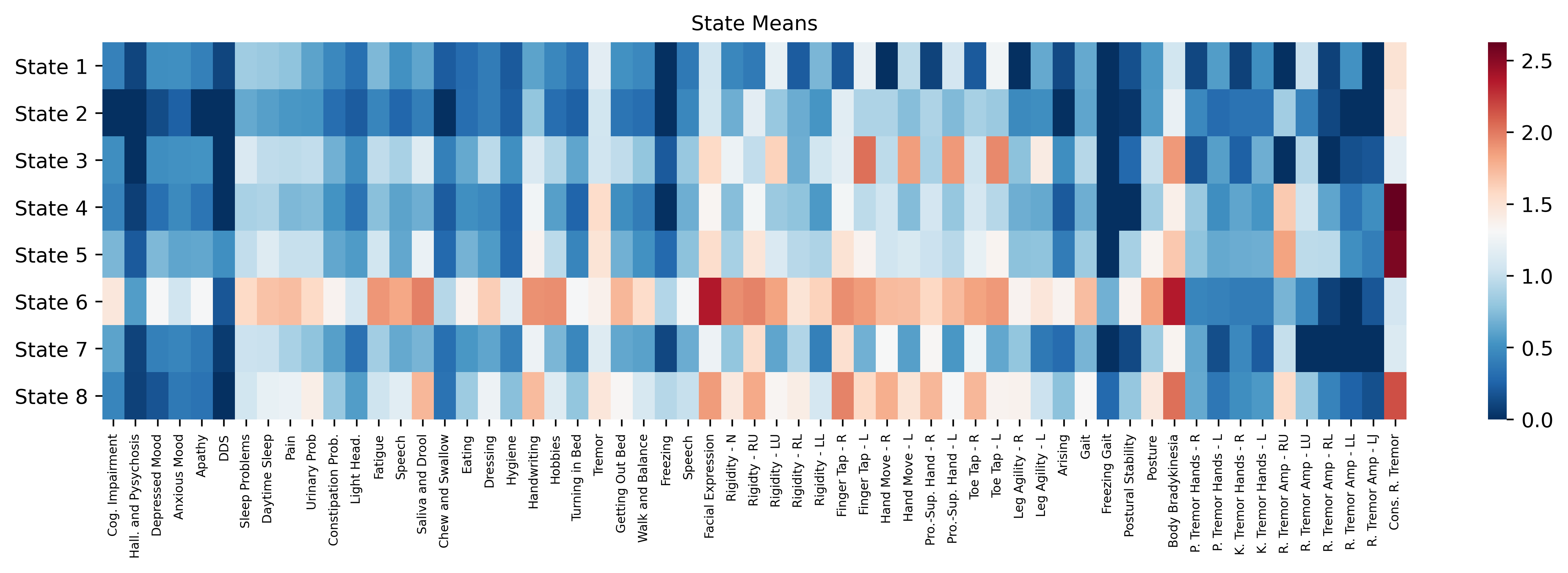}}
\subfigure[mIOHMM]{\label{fig:b}\includegraphics[width=.7\textwidth]{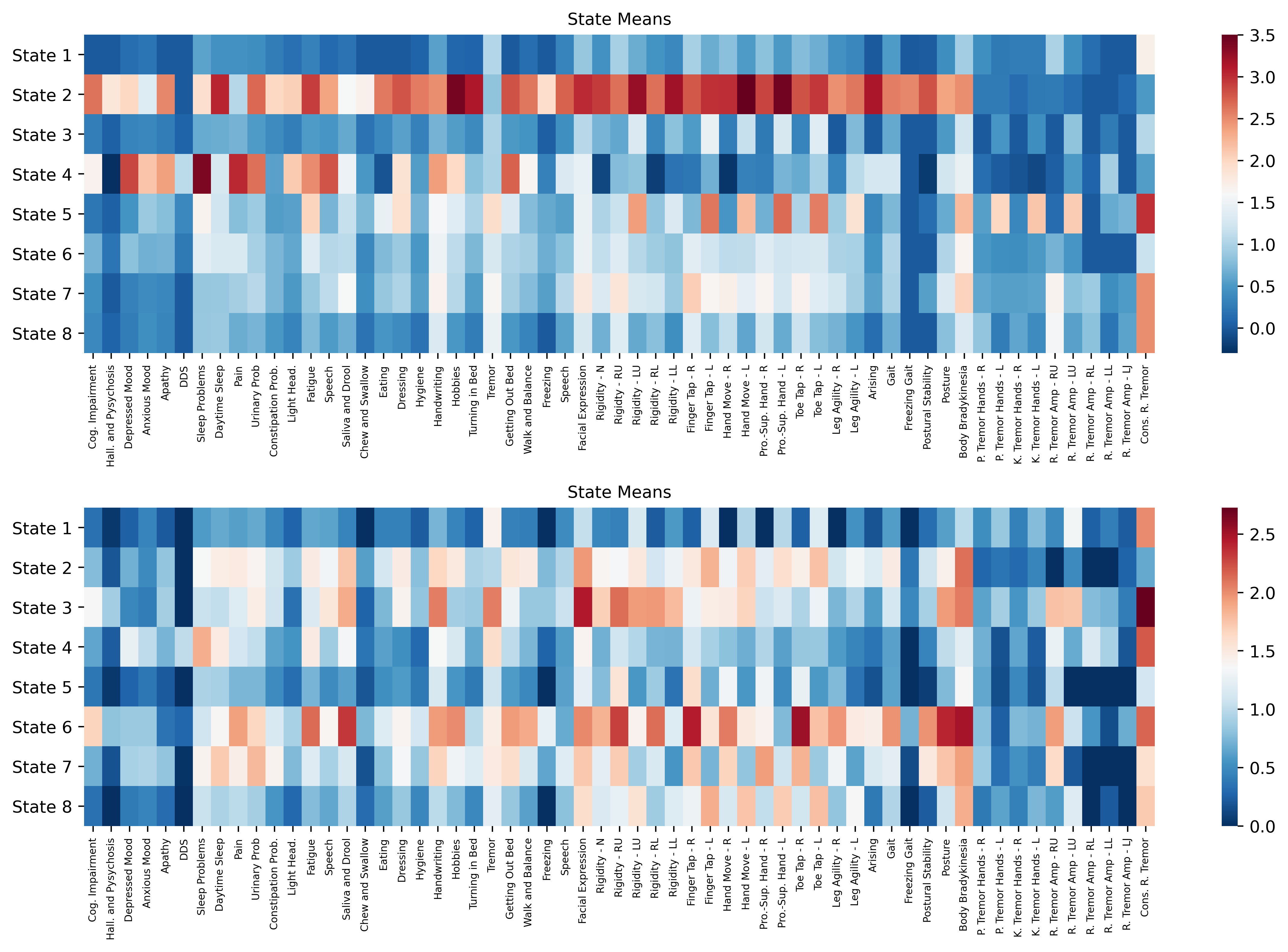}}
\caption{State means obtained using IOHMM and mIOHMMs with 2 mixtures.}\label{fig:appendix:IOHMM-mIOHMM-full-state-means}
\end{figure*}

Fig.\ \ref{fig:appendix:init-tran-params-models} plots the inferred initial-state probabilities and the transition matrices for IOHMM and 2-component mIOHMMs. Fig.\ \ref{fig:appendix:IOHMM-mIOHMM-full-state-means} plots the complete state-means obtained using IOHMM and mIOHMMs with 2 mixtures.

\begin{figure}[htb!]
\begin{center}
\includegraphics[width=.7\textwidth]{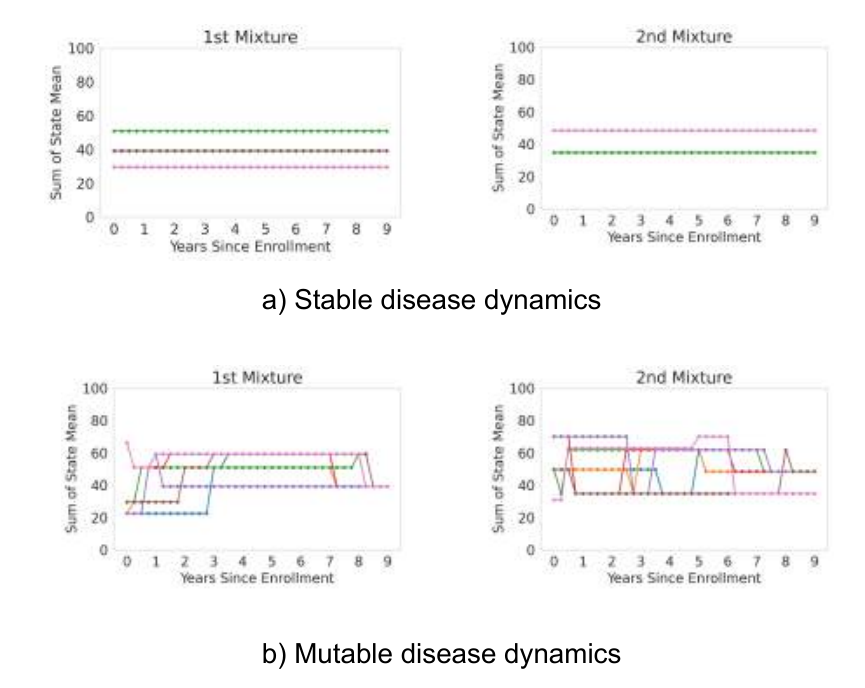}
\caption{State trajectories obtained via the Viterbi algorithm where each line denotes the trajectory for a different patient. Here, X-axis denotes the time index $t^{th}$ for quarterly hospital visits and Y-axis denotes the sum of state-means at the corresponding state. We group trajectories based on the number of states visited. The number of states visited is 1 at Top and 4 at Bottom. }
\label{fig:appendix:state-traj}
\end{center}
\end{figure}

Fig.\ \ref{fig:appendix:state-traj} visualizes the state-trajectories obtained using mIOHMMs for a number of patients and the corresponding overall disease severities, which are calculated based on the estimated states. In particular, we aggregate the state-means over all features. Here, the first and second mixtures are respectively plotted on the left and right hand-side. The figure reflects the intensity difference in the state-means where the second mixture contains more severe symptoms than the first mixture. Moreover, we observe that in the second mixture the final states are more diverse and the jumps are more likely to occur than the first mixture.

\end{document}